\documentclass[letterpaper]{article} 
\usepackage[preprint]{aaai2027}  
\usepackage[hyphens]{url}  
\usepackage{graphicx} 
\usepackage{natbib}  
\usepackage{caption} 
\usepackage{algorithm}
\usepackage{algorithmic}
\usepackage{amsmath}
\usepackage{xcolor}
\usepackage[table]{xcolor}
\usepackage{booktabs}
\usepackage{multirow}
\usepackage{amssymb}
\usepackage{bbding}
\usepackage{subfig}
\usepackage{pifont}
\usepackage{makecell}
\definecolor{ggray}{RGB}{120,120,120}

\definecolor{bestbg}{RGB}{198,234,212}   
\definecolor{secondbg}{RGB}{231,242,255} 
\definecolor{secondbg2}{RGB}{255, 218, 224}
\newcommand{\cmark}{{\ding{51}}}
\newcommand{\xmark}{{\ding{55}}}
\newcommand{\best}[1]{\textbf{#1}}
\newcommand{\second}[1]{\underline{#1}}
\newcommand{\na}{--}
\newcommand{\boldparagraph}[1]{\noindent\textbf{#1}\ }
\usepackage{newfloat}
\usepackage{listings}
\DeclareCaptionStyle{ruled}{labelfont=normalfont,labelsep=colon,strut=off} 
\floatstyle{ruled}
\newfloat{listing}{tb}{lst}{}
\floatname{listing}{Listing}

\usepackage{booktabs}

\title{Focus When Necessary: Adaptive Routing and Collaborative Grounding\\[0.3em] for Training-Free Visual Grounding}
\author{
    Yifan Wang\textsuperscript{\rm 1,\rm 2}, Peiming Li\textsuperscript{\rm 1,\rm 3}, Shiyu Li\textsuperscript{\rm 1}, Zhiyuan Hu\textsuperscript{\rm 1,\rm 3}, Xiaochen Yang\textsuperscript{\rm 4},\\
    Wenming Yang\textsuperscript{\rm 2,}\corresponding, Yang Tang\textsuperscript{\rm 1,}\corresponding\textsuperscript{,}\thanks{Project Lead.}, Zheng Wei\textsuperscript{\rm 1,}\corresponding
}
\affiliations{
    \textsuperscript{\rm 1}Tencent BAC
    \textsuperscript{\rm 2}Shenzhen International Graduate School, Tsinghua University\\
    \textsuperscript{\rm 3}School of Electronic and Computer Engineering, Peking University\\
    \textsuperscript{\rm 4}School of Mathematics and Statistics, University of Glasgow

    \{wyattyfwang, ethanntang, hemingwei\}@tencent.com, yang.wenming@sz.tsinghua.edu.cn
}

\begin{document}

\maketitle

\begin{abstract}
While Multimodal Large Language Models (MLLMs) excel in cross-modal reasoning, they often struggle to perceive fine-grained details in complex high-resolution images. 
Recent training-free methods address this through image scaling and localized cropping. 
However, applying these manipulations indiscriminately introduces computational redundancy for simple queries and can degrade accuracy by truncating essential global context or introducing irrelevant background noise.
To this end, we propose \textbf{LazyMCoT}, a dynamic and training-free framework that adaptively allocates visual grounding efforts based on sample difficulty. 
The framework features an \textbf{\textit{Adaptive Routing}} mechanism that evaluates predictive uncertainty using first-token statistics from a single forward pass. 
This efficiently bypasses confident cases while ensuring the recall of difficult samples via conformal calibration. 
For these challenging cases, a \textbf{\textit{Collaborative Grounding}} module integrates the inherent cross-modal attention of the model with an external visual expert through a two-stage refinement process.
This refinement process generates a precise localized display to recover small or occluded targets. Extensive experiments across diverse benchmarks demonstrate that LazyMCoT rivals training-based approaches by simultaneously improving reasoning accuracy and reducing average inference latency.
Our code is availble at \url{https://github.com/TencentBAC/LazyMCoT}.
\end{abstract}

\section{Introduction}
\label{intro}
Multimodal Large Language Models (MLLMs) have recently achieved unprecedented success across vision-language tasks~\cite{blip-2, instructblip, llava, qwen-vl}. 
While early models with fixed resolution encoders struggle to capture fine details in complex images~\cite{blip-2, llama-adapter, improved}, recent advancements resolve this limitation through dynamic resolution mechanisms~\cite{internvl3, qwen2_5vl, internvl3_5, qwen3vl, llavanext}. 
Building on these robust base models, training-free visual grounding methods have emerged as a promising paradigm~\cite{rap, zoomeye, dyfo}. 
By utilizing visual experts, search algorithms, or attention decoupling, these methods extract key patches and perform localized cropping to enhance the model's perception of subtle visual evidence without the need for expensive retraining~\cite{deepscan, hide, vicrop, look-twice}.

\begin{figure}[t]
   \centering
   \includegraphics[width=1.0\linewidth]{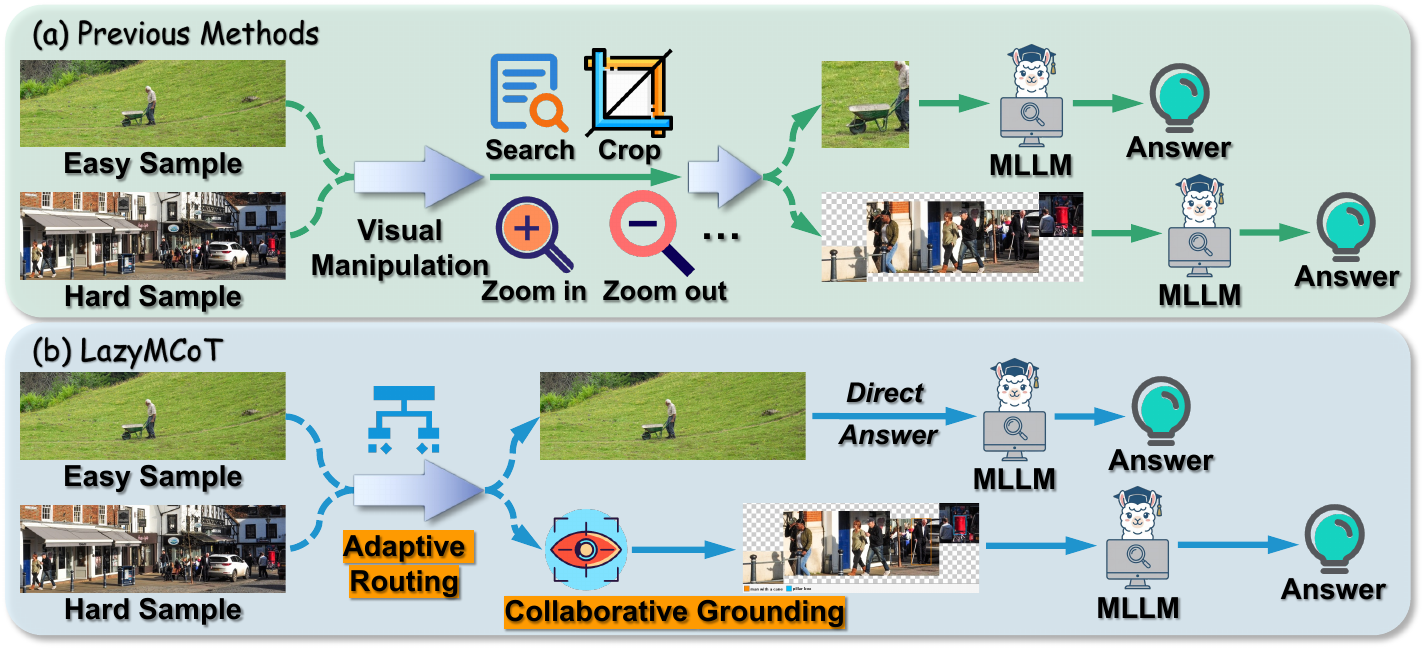}
   \caption{\textbf{LazyMCoT allocates visual grounding effort by sample difficulty.} (a) Previous training-free methods indiscriminately apply heavy visual manipulation to every sample. (b) LazyMCoT instead routes easy samples to a direct answer and dispatches only hard samples through Collaborative Grounding before re-querying the MLLM.}
   \vspace{-8pt}
   \label{fig:intro}
\end{figure}

Although effective on challenging instances, current visual grounding methods without additional training exhibit a critical flaw by indiscriminately applying complex visual manipulations to all samples~\cite{zoomeye, dyfo, gropl2026entropy}. 
Empirical observations reveal that this uniform strategy is highly suboptimal. 
As shown in Fig.~\ref{fig:compare} and Fig.~\ref{fig:compare2}, it wastes significant computational resources since standard VLMs can accurately answer a large fraction of queries in a single forward pass. More importantly, forcibly applying localized cropping to straightforward samples can severely degrade performance. By truncating essential global context and introducing irrelevant background noise, redundant visual processing misguides the model and causes failures on otherwise solvable cases. 
This issue is particularly detrimental in tasks demanding complex reasoning, where blind grounding reduces the overall accuracy of VLMs.

Motivated by these observations, we argue that visual grounding should be applied selectively. We hypothesize that the base VLM's initial predictive uncertainty is a strong indicator of whether further visual exploration is required. Through statistical analysis, we discover that zero-cost features extracted from the first answer token's logits, namely the option top probability and the option-versus-non-option logit gap, exhibit a strong monotonic correlation with the model's predictive entropy. These statistics cleanly separate the samples that the base VLM can solve directly from those that genuinely require dense visual grounding.

To translate this insight into an actionable solution, we propose \textbf{LazyMCoT}, a novel training-free framework that dynamically allocates visual grounding effort based on sample difficulty. LazyMCoT consists of two core components, \textbf{Adaptive Routing} and \textbf{Collaborative Grounding}. 
The Adaptive Routing employs a lightweight decision model to evaluate the first-token statistics, with its decision threshold calibrated via conformal prediction~\cite{conformal_tutorial} to provide a controllable lower bound on the recall of difficult samples. If the model is confident, the router instantly returns the direct answer and bypasses any extra computation. 
Otherwise, the query is routed to the Collaborative Grounding module. 
As shown in Fig.~\ref{fig:intro}, unlike previous methods that rely on inflexible search strategies~\cite{zoomeye, dyfo}, our grounding module couples an attention-driven branch derived from the VLM's cross-modal attention with an external visual expert~\cite{sam3, grounding_dino} branch. Through a two-stage parallel detection and refinement process, it precisely extracts fine-grained evidence containing key targets, and renders a Localized Panel Display, which is then fed back to the VLM for reasoning.

We conduct extensive experiments on multiple challenging benchmarks using open-source VLM backbones. LazyMCoT consistently outperforms existing training-free methods and even matches or exceeds recent training-based grounding approaches. It not only achieves significant accuracy gains on fine-grained localization tasks, but also prevents the performance degradation commonly observed in other grounding methods on reasoning-heavy tasks. Furthermore, by short-circuiting easy samples, LazyMCoT significantly reduces the average end-to-end inference latency.

In summary, our main contributions are threefold:
\begin{itemize}
    \item We identify and empirically validate the inherent limitations of indiscriminate visual grounding in current training-free methods, demonstrating that redundant visual manipulation may hurt performance on easy samples and wastes computational resources.
    \item We propose \textbf{LazyMCoT}, a dynamic framework featuring an \textbf{Adaptive Routing} that leverages zero-cost first-token statistics with conformal-calibrated decision rules to selectively trigger visual grounding, and a \textbf{Collaborative Grounding} module that synergizes VLM attention with visual experts for precise evidence extraction.
    \item Extensive experiments demonstrate that LazyMCoT achieves competitive performance among training-free methods across multiple benchmarks and VLMs, improving accuracy and reducing inference latency.
\end{itemize}

\section{Related Work}
\label{related_work}
\boldparagraph{Multimodal Large Language Models.}
Multimodal Large Language Models (MLLMs) have demonstrated remarkable potential in cross-modal tasks. 
Among these, Vision-Language Models (VLMs) have advanced with particular rapidity.
However, early models~\cite{blip-2, instructblip, llava, qwen-vl, improved} relying on Q-Former~\cite{blip-2} or trainable visual adapters~\cite{llama-adapter} often lose fine-grained details due to their fixed-resolution visual encoders. 
To address complex visual reasoning, dynamic resolution MLLMs~\cite{internvl3, qwen2_5vl, Mini-internvl, expanding, llavanext} have emerged to preserve spatial precision for high-resolution inputs. 
For instance, InternVL-3.5~\cite{internvl3_5} utilizes dynamic slicing for high resolution images, while Qwen3-VL~\cite{qwen3vl} adopts DeepStack multi-layer visual injection for enhanced alignment. 
Building upon these advancements, we propose a training-free method that leverages the inherent capabilities of VLMs to enhance their perception and expand their performance boundaries.

\boldparagraph{Training-Free Visual Grounding.}
Globally uniform feature extraction often wastes computational resources and obscures critical details within background noise. 
Consequently, recent studies emphasize training-free visual grounding and dynamic focusing mechanisms. 
For example, RAP~\cite{rap} pioneers spatially aware layouts, while ZoomEye~\cite{zoomeye} and DyFo~\cite{dyfo} utilize search algorithms for efficient visual navigation. 
Other approaches leverage visual experts for bottom-up evidence extraction~\cite{deepscan}, entropy gradients for region guidance~\cite{gropl2026entropy}, or attention decoupling for key patch extraction~\cite{hide, vicrop, look-twice}. 
Unlike previous methods relying on inflexible search strategies, our proposed approach integrates attention guidance with visual experts to extract fine-grained evidence much more precisely. Additionally, we introduce a dynamic routing mechanism to adaptively schedule inference paths based on varying sample difficulty.

\section{Preliminary}
\label{pre}
\subsection{Limitations of Existing Training-Free Visual Grounding Methods}
\label{limitation_existing_method}
\begin{figure}[t]
   \centering
   \includegraphics[width=0.9\linewidth]{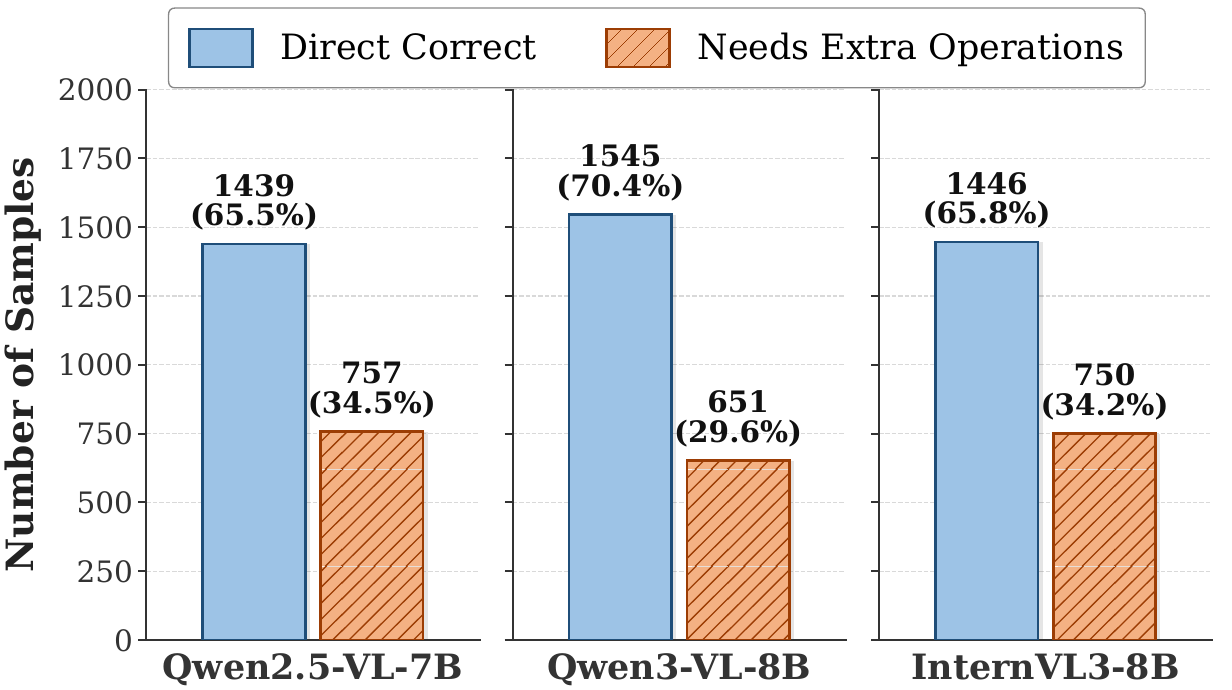}
   \caption{\textbf{Vanilla VLM inference successfully solves most samples.} Breakdown of samples solved without extra visual operations (\textbf{Direct Correct}, blue) vs. those requiring additional grounding (\textbf{Needs Extra Operations}, orange) for each VLM across four benchmarks.}
   \vspace{-8pt}
   \label{fig:compare}
\end{figure}

Recent advancements~\cite{deepscan, hide} in training-free visual grounding heavily rely on image scaling and localized cropping to enhance the perception of fine-grained details. 
However, observations reveal that these operations are not universally beneficial across all instances. 
A substantial portion of straightforward samples can be accurately resolved by original VLMs without additional visual manipulation.
We evaluated on a unified benchmark comprising V* Bench~\cite{V-Star}, HR-Bench 4K/8K~\cite{HR-Bench}, and TreeBench~\cite{TreeBench}. As shown in Fig.~\ref{fig:compare}, an average of \textbf{~67.17\%} of the samples can be correctly answered relying solely on the vanilla VLM inference.

\begin{figure*}[t]
   \centering
   \includegraphics[width=1.0\linewidth]{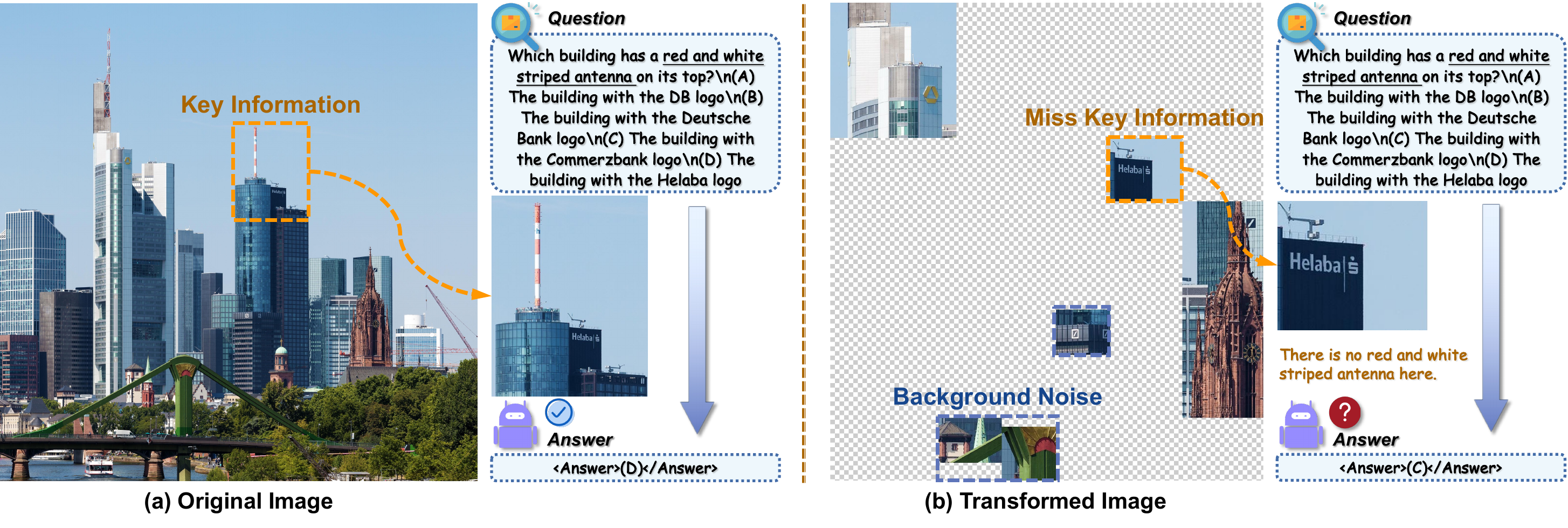}
   \caption{\textbf{Indiscriminately applied visual grounding can hurt easy samples.} (a) On the original image, the base VLM directly localizes the key region and answers correctly. (b) After image scaling and localized cropping, the truncated context omits the key information while introducing background noise, misleading the model into a wrong answer.}
   \label{fig:compare2}
\end{figure*}

Furthermore, forcibly applying these complex operations to such simple cases can introduce unnecessary background noise or truncate essential global context. 
As shown in Fig.~\ref{fig:compare2}, this redundant processing occasionally misguides the model and leads to incorrect predictions for samples that the base VLM would have otherwise answered correctly.

\subsection{Statistical Features for Sample Routing}
\label{sec:stat_features}
Motivated by these observations, we propose selective visual grounding based on sample difficulty. We hypothesize that the base VLM's initial predictive uncertainty indicates the need for further visual exploration. To verify this, we extract two zero-cost statistics from a single forward pass.

Let $ \mathbf{z}\in\mathbb{R}^{V} $ denote the first answer token's logits, and $ \mathcal{O} $ be the candidate option indices. Given the full vocabulary distribution $p_i=\mathrm{softmax}(\mathbf{z})_i$ and the renormalized option-restricted distribution $ \tilde{p} $, we define the option top probability ($\mathrm{topp}$) and option-versus-non-option logit gap ($\Delta_{\mathrm{logit}}$) as:
\begin{equation}
    \mathrm{topp} = \max_{i\in\mathcal{O}} \tilde{p}_i, \quad \Delta_{\mathrm{logit}} = \max_{i\in\mathcal{O}} z_i - \max_{j\notin\mathcal{O}} z_j.
\end{equation}
Intuitively, $\mathrm{topp}$ measures probability concentration on the top option, while $\Delta_{\mathrm{logit}}$ reflects the model's confidence in choosing a valid option over other tokens.
To obtain a comparable scalar, we train a Gradient Boosting Decision Tree~\cite{gbdt} $g_\theta$ on a held-out set to classify $\mathbf{x}=(\mathrm{topp},\,\Delta_{\mathrm{logit}})$ as ori-correct ($y=0$) or ori-wrong ($y=1$). The routing score is the logit of the predicted ori-wrong probability $\hat p(\mathbf{x})=g_\theta(\mathbf{x})$:
\begin{equation}
s(x) \;=\; \log\!\frac{\hat p(\mathbf{x})}{1-\hat p(\mathbf{x})}.
\label{eq:router_score}
\end{equation}

\begin{figure}[t]
    \centering
    \resizebox{1.0\linewidth}{!}{%
        \subfloat[\large Routing score distribution]{
            \includegraphics[height=4.5cm]{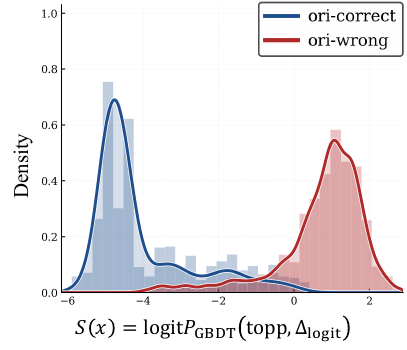}%
            \label{fig:stat_features_a}
        }
        \subfloat[\large Score versus answer entropy]{
            \includegraphics[height=4.5cm]{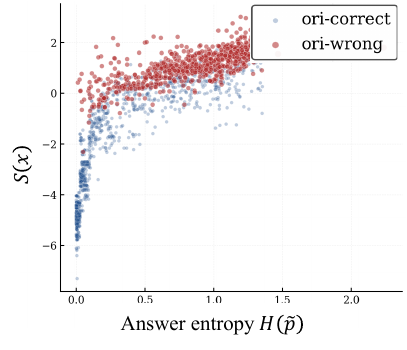}%
            \label{fig:stat_features_b}
        }%
    }
    \caption{\textbf{Statistical separability of ori-correct \& ori-wrong samples on Qwen2.5-VL-7B.} (a) First-token routing score $s(x)$ shows distinct modes per class. (b) $s(x)$ correlates monotonically with answer entropy $H(\tilde p)$, with ori-wrong samples clustering in the high-entropy region.}
    \label{fig:stat_features}
\end{figure}

Using 1,000 samples from Sec.~3.1~\ref{pre} mentioned dataset, we compute option entropy $ H(\tilde{p})=-\sum_{i\in\mathcal{O}} \tilde p_i \log \tilde p_i $ and normalized vocabulary entropy $ \tilde H_{\mathrm{vocab}} = -\sum_{i=1}^{V} p_i \log p_i / \log V $. 
As shown in Fig.~\ref{fig:stat_features}, routing scores separate ori-correct and ori-wrong samples (Fig.~\ref{fig:stat_features_a}), validating $ \mathrm{topp} $ and $ \Delta_{\mathrm{logit}} $ as reliability indicators.
Predictive entropy correlates monotonically with routing scores (Fig.~\ref{fig:stat_features_b}), peaking for ori-wrong samples. 
Thus, these statistics are ideal for a lightweight router to selectively trigger visual grounding.
Please refer to Appendix Sec.~C for these statistics on additional VLMs.

\section{Method}
\label{method}

\subsection{Overview}
\label{sec:overview}
\begin{figure*}[t]
   \centering
   \includegraphics[width=1.0\linewidth]{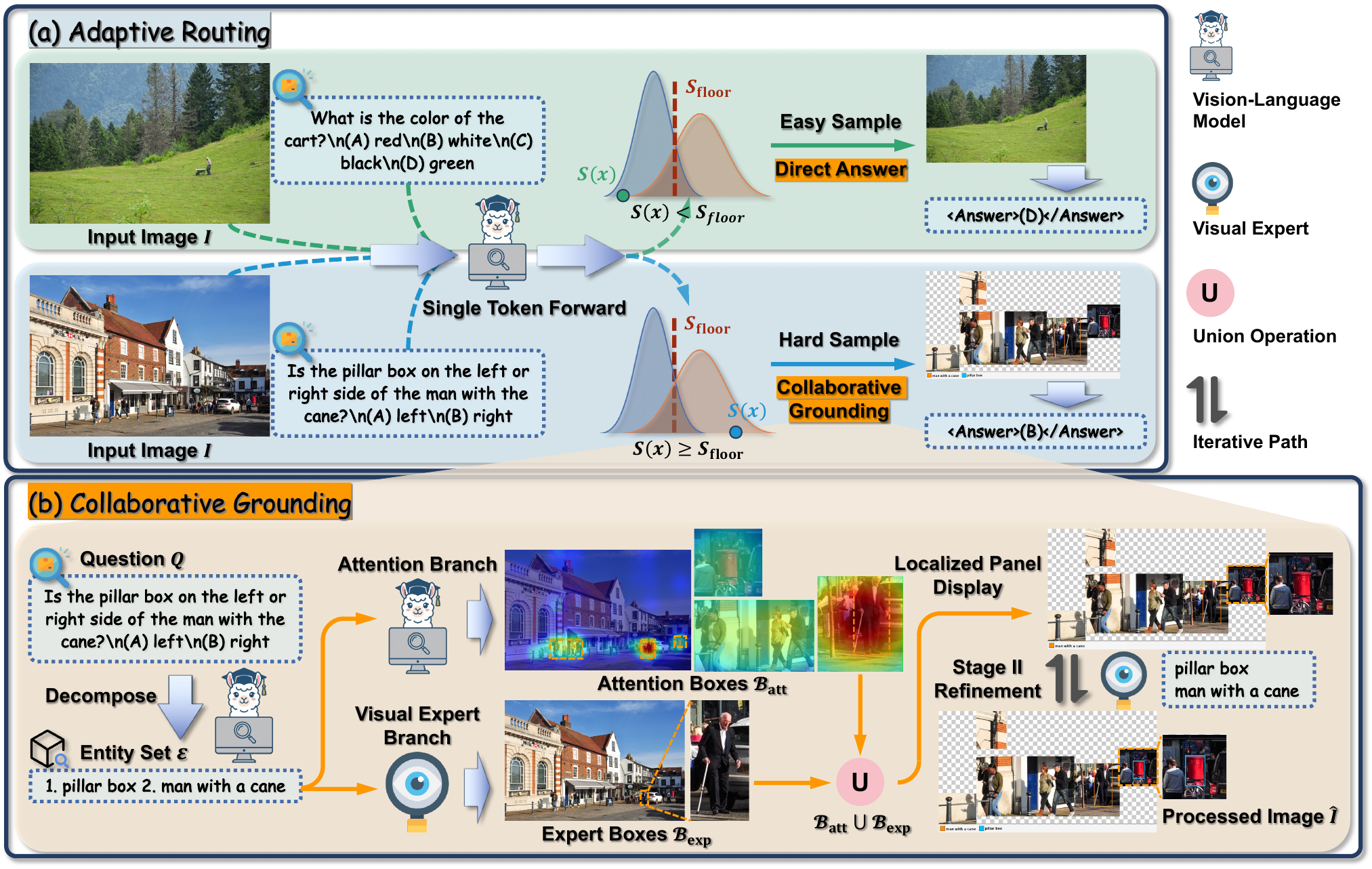}
   \caption{\textbf{Overview of the proposed LazyMCoT framework.} (a) The Adaptive Routing utilizes first-token statistics from a single forward pass to dynamically bypass simple cases or route hard samples. (b) Collaborative Grounding integrates an attention branch and a visual expert to construct a localized panel display for precise VLM re-querying.}
   \vspace{-8pt}
   \label{fig:pipeline}
\end{figure*}

Building on the observations in Sec.~3~\ref{pre}, we propose \textbf{LazyMCoT}, a training-free framework that dynamically allocates visual grounding effort based on sample difficulty. 
As illustrated in Fig.~\ref{fig:pipeline}, given an image $I$ and a multiple-choice question $Q$.
First, a single-token forward pass yields a direct answer and statistics $\mathbf{x}=(\mathrm{topp},\,\Delta_{\mathrm{logit}})$. 
An \textit{Adaptive Routing} module evaluates $\mathbf{x}$ and returns the direct answer immediately if the routing score $s(x) < s_{\mathrm{floor}}$. 
Otherwise, a \textit{Collaborative Grounding} module coupling attention and visual expert branches generates a Localized Panel Display (LPD)~\cite{hide} via a two-stage detection. 
This LPD is then used to re-query the VLM to obtain the final answer. 
This design selectively applies dense grounding to hard samples while preserving zero-shot efficiency on easy ones.

\subsection{Adaptive Routing}
\label{sec:adaptive_routing}
The motivation of adaptive routing is to convert the empirical observations in Sec.~3.2~\ref{sec:stat_features} into an automatic decision rule that triggers Collaborative Grounding only when the base VLM is uncertain. 
Given the first-token feature vector $\mathbf{x}$, we obtain the routing score $s(x)$ in Eqn.~\ref{eq:router_score} via a Gradient Boosting Decision Tree~\cite{gbdt} that is trained once on a held-out routing set $\mathcal{D}_{\mathrm{cal}}$ with labels $y\in\{0,1\}$ for ori-correct and ori-wrong respectively. 
Both $\mathcal{D}_{\mathrm{cal}}$ and the test set are disjoint and the GBDT remains fixed during inference, so the router introduces no extra learnable parameters at deployment.

\boldparagraph{Conformal threshold calibration.}
A naive choice of decision threshold may either skip too many ori-wrong samples or run grounding on too many ori-correct ones. To make the routing behavior controllable, we adopt a conformal calibration on the must-recall subset $\mathcal{D}_{\mathrm{mr}}\subseteq\mathcal{D}_{\mathrm{cal}}$, containing all ori-wrong samples benefiting from grounding. Let $\{s_i\}_{i\in\mathcal{D}_{\mathrm{mr}}}$ be the out-of-fold scores produced by the GBDT. For a target miscoverage rate $\alpha\in[0,1)$, the routing threshold is set to:
\begin{equation}
s_{\mathrm{floor}} \;=\; Q_{\alpha}\!\bigl(\{s_i\}_{i\in\mathcal{D}_{\mathrm{mr}}}\bigr),
\label{eq:s_floor}
\end{equation}
where $Q_{\alpha}(\cdot)$ denotes the empirical $\alpha$-quantile. By construction, at most an $\alpha$ fraction of must-recall samples falls below $s_{\mathrm{floor}}$, guaranteeing a controlled lower bound on the recall of difficult samples. Smaller $\alpha$ yields a lower threshold and a more conservative router that triggers grounding more often, while larger $\alpha$ allows aggressive skipping of easy samples.

\boldparagraph{Routing rule.}
Let $\mathrm{Direct}(I,Q)$ denote the direct answer obtained from the single-token forward pass and let $\mathrm{CG}(I,Q)$ denote the answer produced by feeding the LPD $\hat I$ back to the VLM. The final prediction of LazyMCoT is:
\begin{equation}
\hat y \;=\;
\begin{cases}
\mathrm{Direct}(I,Q), & s(x) < s_{\mathrm{floor}},\\[2pt]
\mathrm{CG}(I,Q), & s(x) \ge s_{\mathrm{floor}}.
\end{cases}
\label{eq:routing_rule}
\end{equation}
Because $s(x)$ is computed from the same forward pass as the direct answer and $\mathcal{D}_{\mathrm{cal}}$ is routing-data only, no additional inference cost is introduced for skipped samples and no training of the base VLM is required. This makes adaptive routing a lightweight plug-in that can be paired with any VLM.

\subsection{Collaborative Grounding}
\label{sec:collab_grounding}
\boldparagraph{Entity decomposition and parallel detection.}
For routed difficult samples, we first prompt the VLM with a rule-based template~\cite{hide} to decompose $Q$ into a list of canonical entities $\mathcal{E}=\{e_1,\dots,e_M\}$. With $\mathcal{E}$ as queries, we run two complementary detectors in parallel. 
The \emph{visual expert branch} feeds $\mathcal{E}$ into SAM3~\cite{sam3} to obtain a set of expert boxes $\mathcal{B}_{\mathrm{exp}}$. The \emph{attention branch} appends the prompt ``Search the following entities in the images: $\mathcal{E}$'' to the VLM input and records the cross-modal attention $A\in\mathbb{R}^{T\times N}$, where $T$ is the number of entity tokens and $N$ is the number of visual tokens. 
Per entity token, the attention map is reshaped to the spatial grid, smoothed by a Gaussian kernel of bandwidth $\sigma$, normalized, and aggregated across entity tokens into a single saliency map:
\begin{equation}
\mathcal{A}(I) \;=\; \frac{1}{T}\sum_{t=1}^{T}\,\mathcal{N}\!\bigl(g_\sigma * A_t\bigr),
\label{eq:agg_attn}
\end{equation}
where $\mathcal{N}(\cdot)$ rescales the map to $[0,1]$. Connected components of the saliency map above a relative threshold $\tau$ on $\mathcal{A}(I)$ are converted into the attention boxes $\mathcal{B}_{\mathrm{att}}$.

\boldparagraph{Two-stage refinement.}
We observe that $\mathcal{B}_{\mathrm{exp}}$ tends to recall the most salient instances but may miss small or occluded ones, whereas $\mathcal{B}_{\mathrm{att}}$ covers question-relevant regions but is noisy. To exploit their complementarity, we couple the two sources by a two-stage detection procedure. In the first stage, we take the union $\mathcal{B}^{(1)} = \mathcal{B}_{\mathrm{att}} \cup \mathcal{B}_{\mathrm{exp}}$ as a coarse evidence pool. In the second stage, for each box $b\in\mathcal{B}^{(1)}_{\mathrm{att}}$ that is not already covered by $\mathcal{B}_{\mathrm{exp}}$, we crop $I$ to the slightly enlarged region of $b$ and re-query SAM3 with the same entities. The newly discovered boxes $\Delta\mathcal{B}$ inside the crop are mapped back to image coordinates and appended to the evidence pool. The refined box set $\mathcal{B}^{(2)} = \mathcal{B}^{(1)} \cup \Delta\mathcal{B}$ is finally rendered as a localized panel display $\hat I$, in which each region is assigned a color and a textual legend so that the VLM can reason over multiple evidence patches in one forward pass. The complete procedure is summarized in Alg.~\ref{alg:collab_grounding}.

\begin{algorithm}[tb]
\caption{Collaborative Grounding}
\label{alg:collab_grounding}
\textbf{Input}: image $I$, question $Q$, base VLM $f$, visual expert $\mathcal{S}$\\
\textbf{Parameter}: kernel bandwidth $\sigma$, threshold $\tau$\\
\textbf{Output}: localized panel display $\hat I$
\begin{algorithmic}[1]
\STATE Decompose $Q$ into entity set $\mathcal{E}=\{e_1,\dots,e_M\}$ via $f$.
\STATE $\mathcal{B}_{\mathrm{exp}} \leftarrow \mathcal{S}(I,\mathcal{E})$ \quad\textit{(visual expert branch)}
\STATE Run $f$ with prompt ``Search $\mathcal{E}$ in $I$'' and record cross-modal attention $A$.
\STATE Compute aggregated saliency $\mathcal{A}(I)$ by Eqn.~\ref{eq:agg_attn}.
\STATE Threshold $\mathcal{A}(I)$ at $\tau$ to obtain attention boxes $\mathcal{B}_{\mathrm{att}}$.
\STATE $\mathcal{B}^{(1)} \leftarrow \mathcal{B}_{\mathrm{att}} \cup \mathcal{B}_{\mathrm{exp}}$.
\STATE $\Delta\mathcal{B} \leftarrow \emptyset$
\FOR{each $b \in \mathcal{B}_{\mathrm{att}}$ not covered by $\mathcal{B}_{\mathrm{exp}}$}
    \STATE $I_b \leftarrow$ crop of $I$ on the enlarged region of $b$.
    \STATE $\Delta\mathcal{B} \leftarrow \Delta\mathcal{B} \cup \mathcal{S}(I_b,\mathcal{E})$.
\ENDFOR
\STATE $\mathcal{B}^{(2)} \leftarrow \mathcal{B}^{(1)} \cup \Delta\mathcal{B}$
\STATE Render $\mathcal{B}^{(2)}$ on $I$ with color borders and legends to obtain $\hat I$.
\STATE \textbf{return} $\hat I$
\end{algorithmic}
\end{algorithm}

\section{Experiments}
\label{exp}
\subsection{Experiment Settings}
\label{experiment settings}
\boldparagraph{Datasets \& Metrics.}
We systematically evaluate our proposed method on three challenging benchmarks. 
(1) \textbf{V* Bench}~\cite{V-Star}, which contains 191 images with an average resolution of 2246$\times$1582 and focuses on \textit{Direct Attribute} recognition (Att.) and \textit{Spatial Relationship} reasoning (Spa.).
(2) \textbf{HR-Bench}~\cite{HR-Bench}, which provides 4K and 8K resolution images that are evenly split into \textit{Single-Instance} (Sin.) and \textit{Cross-Instance} (Cro.) perception tasks. 
(3) \textbf{TreeBench}~\cite{TreeBench}, comprising 405 images with an average resolution of 2152$\times$1615, which covers fine-grained perception (e.g., Material Recognition, OCR) and multi-step reasoning (e.g., Occlusion, Comparative Analysis). 
Multiple-choice accuracy is adopted as the primary evaluation metric across all benchmarks.

\boldparagraph{Implementation Details.}
We employ SAM3~\cite{sam3} as the visual expert, setting the maximum number of detected objects to $k=10$ to balance performance and latency. 
LazyMCoT is evaluated across three different VLMs: Qwen2.5-VL-7B~\cite{qwen2_5vl}, Qwen3-VL-8B~\cite{qwen3vl}, and InternVL3-8B~\cite{internvl3}. 
For the routing strategy, we train a Gradient Boosting Decision Tree (GBDT)~\cite{gbdt} classifier to determine the routing threshold without manual tuning. Conformal prediction is subsequently applied to the out-of-fold scores to ensure high recall for mispredicted samples. 
Please refer to Appendix Sec.~A for further implementation details.

\subsection{Main Results}
\boldparagraph{Results on HR-Bench and V$^\ast$.}
As reported in Tab.~\ref{tab:hr_vstar}, LazyMCoT delivers consistent gains over the base VLMs across all three backbones, raising the average accuracy by $3.3$, $7.9$, and $8.4$ points on InternVL3-8B, Qwen3-VL-8B, and Qwen2.5-VL-7B, respectively. Among training-free methods, LazyMCoT ranks first on every aggregate column and surpasses the HiDe baseline on average. 
Notably, on the Qwen2.5-VL-7B, our training-free framework matches or exceeds recent training-based grounding methods, achieving the best V$^\ast$ average of $90.6\%$ without any parameter update. The improvement is most pronounced on tasks that demand fine-grained spatial localization (V$^\ast$-Spa.~$+17.1$ points on Qwen2.5-VL-7B), confirming that our collaborative grounding effectively recovers small targets.

\begin{table*}
\centering
\fontsize{5pt}{5pt}\selectfont
\renewcommand{\arraystretch}{0.90}
\setlength{\tabcolsep}{5pt}
\resizebox{\textwidth}{!}{
\begin{tabular}{lcccccccccc}                       
\toprule
\multirow{2}{*}[-0.6ex]{\textbf{Method}} & \multirow{2}{*}[-0.6ex]{\makecell{\textbf{Training}\\\textbf{Free}}} 
 & \multicolumn{3}{c}{\textbf{HR-Bench-4K}} & \multicolumn{3}{c}{\textbf{HR-Bench-8K}} & \multicolumn{3}{c}{\textbf{V*}} \\
\cmidrule(lr){3-5} \cmidrule(lr){6-8} \cmidrule(lr){9-11}
 & & Sin. & Cro. & Avg. & Sin. & Cro. & Avg. & Att. & Spa. & Avg. \\
\midrule
GPT-4o~\cite{gpt4o} & \na & 70.00 & 48.00 & 59.00 & 62.00 & 49.00 & 55.50 & \na & \na & 66.00 \\
\midrule
\rowcolor{gray!20}\multicolumn{11}{c}{\textbf{\textit{InternVL3-8B-Instruct}}} \\                 
\noalign{\vspace{2pt}}
Base~\cite{internvl3} & \na & 82.80 & \best{58.80} & 70.80 & 80.00 & \best{59.80} & 69.90 & 81.70 & 78.90 & 80.60 \\
ViCrop~\cite{vicrop} & \cmark & 88.00 & 57.00 & 72.50 & 82.80 & \second{54.80} & 68.80 & 88.70 & 75.00 & 83.30 \\
HiDe~\cite{hide} & \cmark & \second{88.50} & \second{57.50} & \second{73.00} & \second{86.00} & 54.20 & \second{70.10} & \second{89.60} & \second{81.60} & \second{86.40} \\
\textbf{LazyMCoT (Ours)} & \cmark & \best{89.50} & \best{58.80} & \best{74.10} & \best{89.80} & 53.00 & \best{71.40} & \best{89.70} & \best{83.90} & \best{87.30} \\
$\Delta$ \textit{vs.} InternVL3-8B-Instruct & \na & $\uparrow$6.70 & \na & $\uparrow$3.30 & $\uparrow$9.80 & $\downarrow$6.80 & $\uparrow$1.50 & $\uparrow$8.00 & $\uparrow$5.00 & $\uparrow$6.70 \\
\midrule
\rowcolor{gray!20}\multicolumn{11}{c}{\textbf{\textit{Qwen3-VL-8B-Instruct}}} \\                  
\noalign{\vspace{2pt}}
Base~\cite{qwen3vl} & \na & \second{90.50} & \second{66.30} & \second{78.40} & \second{83.80} & \best{65.00} & \second{74.40} & \second{85.20} & \second{80.30} & \second{83.20} \\
\textbf{LazyMCoT (Ours)} & \cmark & \best{94.30} & \best{66.50} & \best{80.40} & \best{90.80} & \second{64.00} & \best{77.40} & \best{92.20} & \best{89.50} & \best{91.10} \\
$\Delta$ \textit{vs.} Qwen3-VL-8B-Instruct & \na & $\uparrow$3.80 & $\uparrow$0.20 & $\uparrow$2.00 & $\uparrow$7.00 & $\downarrow$1.00 & $\uparrow$3.00 & $\uparrow$7.00 & $\uparrow$9.20 & $\uparrow$7.90 \\
\midrule
\rowcolor{gray!20}\multicolumn{11}{c}{\textbf{\textit{Qwen2.5-VL-7B-Instruct}}} \\                
\noalign{\vspace{2pt}}
Base~\cite{qwen2_5vl}   & \na    & 88.80 & 54.80 & 71.80 & 84.20 & 51.50 & 67.90 & 80.90 & 76.30 & 79.10 \\
DeepEyes~\cite{deepeyes}& \xmark & 91.30 & 59.00 & 75.10 & 86.80 & 58.50 & 72.60 & 92.10 & 86.80 & 90.00 \\
Thyme-VL~\cite{thyme}   & \xmark & 91.00 & 63.00 & 77.00 & 86.50 & 57.50 & 72.00 & 83.50 & 80.30 & 82.20 \\
TreeVGR~\cite{TreeBench}& \xmark & 89.50 & 61.50 & 72.70 & 84.40 & 57.20 & 69.80 & 86.10 & 85.50 & 85.90 \\
\cmidrule(lr){1-11}                                                    
ViCrop~\cite{vicrop}        & \cmark & 90.50 & 57.50 & 74.00 & 85.50 & 53.00 & 69.30 & 89.60 & 71.10 & 82.20 \\
Dyfo~\cite{dyfo}            & \cmark & 89.20 & 53.50 & 71.30 & 86.50 & 53.20 & 69.80 & 82.60 & \second{86.80} & 84.30 \\
ZoomRefine~\cite{zoom-refine}& \cmark & 88.50 & 55.30 & 71.50 & 83.90 & 54.00 & 68.60 & 85.30 & 77.60 & 82.20 \\
DeepScan~\cite{deepscan}    & \cmark & 90.10 & \best{59.70} & 75.00 & 87.20 & \best{57.60} & 72.40 & \second{93.00} & \second{86.80} & \best{90.60} \\
HiDe~\cite{hide}            & \cmark & \second{95.50} & \second{59.30} & \second{77.40} & \second{92.50} & \second{57.30} & \second{74.90} & \best{94.80} & 82.90 & \second{90.00} \\
\textbf{LazyMCoT (Ours)}    & \cmark & \best{96.20} & 59.00 & \best{77.60} & \best{93.80} & 57.00 & \best{75.40} & 92.20 & \best{88.20} & \best{90.60} \\
$\Delta$ \textit{vs.} Qwen2.5-VL-7B-Instruct & \na & $\uparrow$5.70 & $\uparrow$1.50 & $\uparrow$3.60 & $\uparrow$8.30 & $\uparrow$4.00 & $\uparrow$6.10 & $\uparrow$2.60 & $\uparrow$17.1 & $\uparrow$8.40 \\
\bottomrule
\end{tabular}
}
\caption{Quantitative results on HR-Bench and V$^\ast$ benchmarks. For each backbone group, the \textbf{best} result is in \textbf{bold} and the \underline{second-best} is \underline{underlined}. LazyMCoT achieves competitive performance across all settings.}
\vspace{-8pt}
\label{tab:hr_vstar}
\end{table*}

\boldparagraph{Results on TreeBench.}
TreeBench probes perception and reasoning beyond pure localization, where indiscriminate grounding can disrupt models with already accurate reasoning chains.
As shown in Tab.~\ref{tab:treebench}, on Qwen3-VL-8B HiDe degrades the average score from $43.0\%$ to $41.7\%$ because every sample is forced into the grounding pipeline, whereas LazyMCoT lifts it to $43.5\%$ by routing only difficult cases through the visual expert. 
On the Qwen2.5-VL-7B backbone, LazyMCoT obtains $41.7\%$ average accuracy with balanced gains over both Perception ($+3.4$ points vs. base) and Reasoning categories, outperforming all training-free competitors. 
These results confirm that adaptive routing is essential for benchmarks where blind grounding is detrimental.

\begin{figure}[t]
   \centering
   \includegraphics[width=1.0\linewidth]{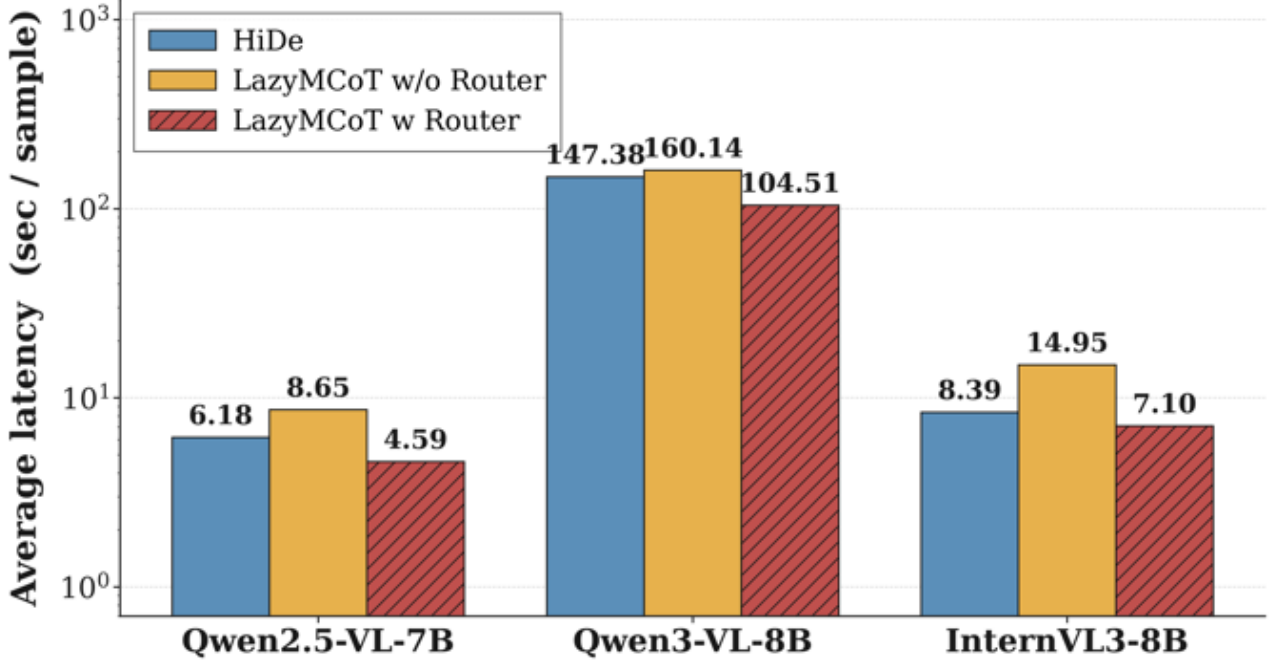}
   \caption{\textbf{Adaptive routing yields the lower end-to-end latency.} Average per-sample inference time on V$^\ast$ for three VLM backbones under three configurations: HiDe, w/o Adaptive Routing, and the full LazyMCoT. The first-token fast-path lets LazyMCoT skip confident samples.}
   \vspace{-8pt}
   \label{fig:time}
\end{figure}

\boldparagraph{Latency comparison.}
Fig.~\ref{fig:time} reports the average per-sample wall-clock time on V* across three VLMs, comparing HiDe, w/o Adaptive Routing, and our full LazyMCoT. 
To ensure fair comparison, all experiments are conducted on a single NVIDIA H20 GPU with a batch size of 1. 
Adding collaborative grounding alone is slightly slower than HiDe due to the SAM3 verification stage, but enabling the router lets the framework short-circuit on samples whose first-token statistics already indicate confident answers, reducing average inference latency.
Hence, LazyMCoT offers a balance of reasoning accuracy and efficiency for visual grounding.

\begin{table*}
\centering
\renewcommand{\arraystretch}{0.8}      
\setlength{\tabcolsep}{4pt}             
\small                            
\resizebox{\textwidth}{!}{
\begin{tabular}{lcccccccccccc}
\toprule
\multirow[c]{2}{*}[-3.0ex]{\textbf{Method}} & \multirow[c]{2}{*}[-3.0ex]{\textbf{Avg.}} 
  & \multicolumn{5}{c}{\textbf{Perception}} 
  & \multicolumn{5}{c}{\textbf{Reasoning}} \\
\cmidrule(lr){3-7} \cmidrule(lr){8-12}
 & & \rotatebox{35}{Attributes} & \rotatebox{35}{Material} & \rotatebox{35}{Phy. State} & \rotatebox{35}{Obj. Retr.} & \rotatebox{35}{OCR} 
   & \rotatebox{35}{Per. Trans.} & \rotatebox{35}{Ordering} & \rotatebox{35}{Con. \& Oc.} & \rotatebox{35}{Spa. Cont.} & \rotatebox{35}{Comparison} \\
\midrule
GPT-4o~{\cite{gpt4o}} & 46.90 & 51.70 & 61.50 & 65.20 & 43.80 & 69.10 & 18.80 & 38.60 & 48.80 & 72.40 & 43.20  \\
\midrule
\rowcolor{gray!20}\multicolumn{12}{c}{\textbf{\textit{Qwen2.5-VL-7B-Instruct}}} \\
\noalign{\vspace{2pt}}
Base~{\cite{qwen2_5vl}} & 37.00 & \second{55.20} & 53.80 & \best{56.50} & \second{62.50} & 27.90 & 20.00 & \best{35.10} & 39.00 & \second{44.80} & 43.20 \\
Dyfo~{\cite{dyfo}} & 39.30 & \best{58.60} & \best{69.20} & \best{56.50} & \second{62.50} & \second{35.30} & \best{21.20} & \best{35.10} & \second{41.50} & \second{44.80} & 40.90 \\
ZoomRefine~{\cite{zoom-refine}} & 38.00 & 48.30 & \second{61.50} & \best{56.50} & \second{62.50} & \best{39.70} & \second{18.80} & 29.80 & \best{46.30} & \second{44.80} & 38.60 \\
HiDe~{\cite{hide}} & \second{40.00} & \second{55.20} & \second{61.50} & \second{52.20} & 56.20 & \best{39.70} & \best{21.20} & \second{31.60} & \best{46.30} & \best{48.30} & \second{47.70} \\
\textbf{LazyMCoT (Ours)} & \best{41.70} & \best{58.60} & \second{61.50} & \best{56.50} & \best{68.80} & \best{39.70} & \best{21.20} & \second{31.60} & \best{46.30} & \best{48.30} & \best{54.50} \\
\midrule
\rowcolor{gray!20}\multicolumn{12}{c}{\textbf{\textit{Qwen3-VL-8B-Instruct}}} \\
\noalign{\vspace{2pt}}
Base~{\cite{qwen3vl}} & \second{43.00} & \best{55.20} & \best{61.50} & \best{69.60} & \best{75.00} & \best{45.60} & \best{18.80} & \second{28.10} & 41.50 & \second{69.00} & \second{50.00} \\
HiDe~{\cite{hide}} & 41.70 & \second{48.30} & \second{46.20} & \second{65.20} & \best{75.00} & \best{45.60} & 15.30 & 24.60 & \best{53.70} & 65.50 & \best{52.30} \\
\textbf{LazyMCoT (Ours)} & \best{43.50} & \best{55.20} & \best{61.50} & \second{65.20} & \best{75.00} & \best{45.60} & \second{16.50} & \best{29.80} & \second{48.80} & \best{72.40} & \second{50.00} \\
\bottomrule
\end{tabular}
}
\caption{Quantitative results on the TreeBench benchmark. The \textbf{best} results for each VLM are highlighted in \textbf{bold}, and the \underline{second-best} is \underline{underline}. Our method effectively retains and improves the perception capability.}
\label{tab:treebench}
\end{table*}

\begin{figure*}[t]
   \centering
   \includegraphics[width=1.0\linewidth]{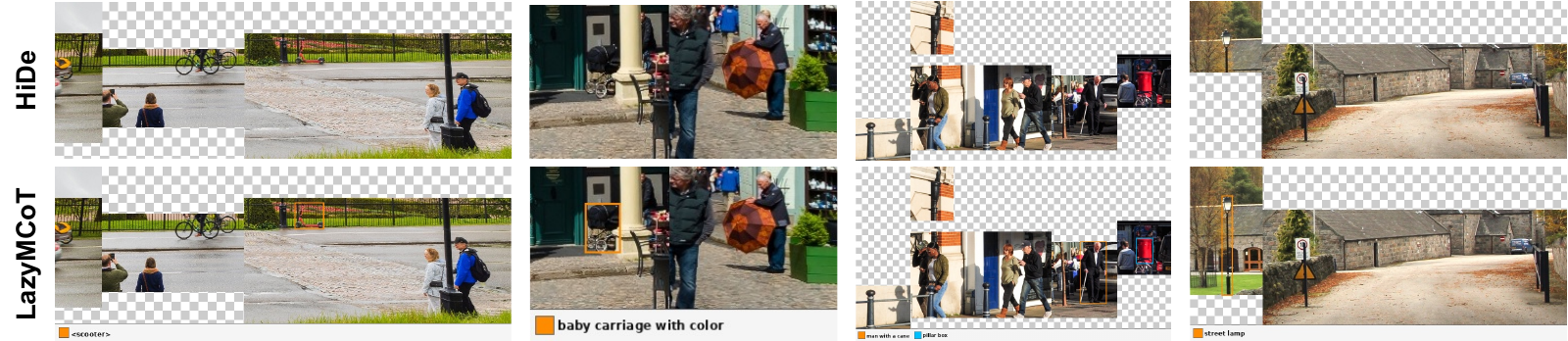}
   \caption{\textbf{Qualitative comparison on hard samples.} HiDe and LazyMCoT results are shown in the top and bottom rows. By recovering small or co-occurring targets missed by HiDe, our method provides more complete evidence for VLM re-querying.}
   \vspace{-8pt}
   \label{fig:case}
\end{figure*}

\subsection{Ablation Study}
\boldparagraph{Effect of the two main components.}
Tab.~\ref{tab:ablation_components} dissects Adaptive Routing (AR) and Collaborative Grounding (CG). 
Forcing CG on every sample lifts V$^\ast$ from $79.1\%$ to $90.0\%$ and HR-Bench-4K from $71.8\%$ to $77.4\%$, but on TreeBench the gain is limited because indiscriminate grounding hurts easy reasoning samples. 
Plugging in AR further pushes V$^\ast$ to $90.6\%$ and TreeBench to $41.7\%$, indicating that the two components are complementary. CG provides the precision needed for hard cases, while AR shields the base VLM from unnecessary grounding on easy ones.

\begin{table}[t]
\centering
\fontsize{8pt}{7pt}\selectfont
\renewcommand{\arraystretch}{1.05}
\setlength{\tabcolsep}{5pt}
\begin{tabular}{cccccc}
\toprule
AR & CG & V$^\ast$ & HR-4K & HR-8K & TreeBench \\
\midrule
\xmark & \xmark & 79.10 & 71.80 & 67.90 & 37.00 \\
\xmark & \cmark & 90.00 & 77.40 & 74.90 & 40.00 \\
\rowcolor{gray!20}\cmark & \cmark & \textbf{90.60} & \textbf{77.60} & \textbf{75.40} & \textbf{41.70} \\
\bottomrule
\end{tabular}
\caption{Ablation study on Adaptive Routing (AR) and Collaborative Grounding (CG) across multiple benchmarks.}
\label{tab:ablation_components}
\end{table}

\boldparagraph{Two-stage visual expert refinement.}
Tab.~\ref{tab:ablation_stages} ablates CG's inner structure. 
The attention branch ($\mathcal{B}_{\mathrm{att}}$) and the visual expert branch ($\mathcal{B}_{\mathrm{exp}}$) alone reach $80.3\%$ and $85.9\%$ on V$^\ast$ respectively, since each has its own failure mode, like noisy attention regions or missed small instances. 
Their Stage~1 union $\mathcal{B}^{(1)}$ lifts V$^\ast$ by $2.5$ points over the stronger single source. 
Adding the Stage~2 refinement, which re-queries SAM3 inside enlarged attention crops to recover small or occluded targets, brings the final V$^\ast$ to $90.6\%$, validating that the two stages are necessary for accurate localization.

\begin{table}[t]
\centering
\fontsize{8pt}{7pt}\selectfont
\renewcommand{\arraystretch}{1.05}
\setlength{\tabcolsep}{4pt}
\begin{tabular}{cccccc}
\toprule
$\mathcal{B}_{\mathrm{att}}$ & $\mathcal{B}_{\mathrm{exp}}$ & Stage~2 & V$^\ast$-Att. & V$^\ast$-Spa. & V$^\ast$-Avg. \\
\midrule
\cmark & \xmark & \xmark & 84.30 & 76.30 & 80.30 \\
\xmark & \cmark & \xmark & 88.70 & 82.90 & 85.90 \\
\cmark & \cmark & \xmark & 91.30 & 85.50 & 88.40 \\
\rowcolor{gray!20}\cmark & \cmark & \cmark & \textbf{92.20} & \textbf{88.20} & \textbf{90.60} \\
\bottomrule
\end{tabular}
\caption{Ablation study on the two-stage Collaborative Grounding pipeline on V* Bench.}
\vspace{-8pt}
\label{tab:ablation_stages}
\end{table}

\boldparagraph{Effect of conformal miscoverage rate $\boldsymbol{\alpha}$.}
Tab.~\ref{tab:ablation_alpha} sweeps the conformal miscoverage rate $\alpha$ that controls $s_{\mathrm{floor}}$ via Eqn.~\ref{eq:s_floor}. 
A smaller $\alpha$ yields a lower threshold and a more conservative router that triggers grounding on most samples, while a larger $\alpha$ aggressively skips and falls back to the base VLM. 
The sweet spot at $\alpha=0$ achieves the best V$^\ast$ of $90.6\%$ while sparing $24.7\%$ samples from the grounding pipeline, showing that conformal calibration offers a principled and tunable trade-off between accuracy and efficiency.

\begin{table}
\centering
\fontsize{8pt}{7pt}\selectfont
\renewcommand{\arraystretch}{1.05}
\setlength{\tabcolsep}{6pt}
\begin{tabular}{cccccc}
\toprule
$\alpha$ & $s_{\mathrm{floor}}$ & Skip Rate (\%) & V$^\ast$-Att. & V$^\ast$-Spa. & V$^\ast$-Avg. \\
\midrule
\rowcolor{gray!20}{0.00} & ${-0.25}$ & {24.7} & {92.20} & \textbf{88.20} & \textbf{90.60} \\
0.05 & $-0.74$ & 51.8 & 93.30 & 86.20 & 90.20 \\
0.10 & $-4.61$ & 58.1 & \textbf{94.10} & 82.40 & 89.25 \\
0.15 & $\phantom{-}0.01$ & 62.3 & 91.40 & 87.30 & 89.85 \\
0.20 & $\phantom{-}0.20$ & 64.9 & 89.10 & 83.20 & 86.65 \\
0.30 & $\phantom{-}0.47$ & 71.2 & 86.20 & 79.60 & 83.45 \\
0.50 & $\phantom{-}0.94$ & 84.8 & 82.80 & 77.10 & 80.45 \\
0.70 & $\phantom{-}1.29$ & \textbf{94.2} & 81.10 & 76.50 & 79.25 \\
\bottomrule
\end{tabular}
\caption{Ablation on the conformal miscoverage rate $\alpha$. Results on V$^\ast$ Bench with Qwen2.5-VL-7B.}
\vspace{-8pt}
\label{tab:ablation_alpha}
\end{table}

\subsection{Case Study}
Fig.~\ref{fig:case} presents qualitative comparisons of the LPD produced by HiDe and our LazyMCoT on four representative hard samples from V$^\ast$ and HR-Bench.
HiDe relies on attention-driven cropping and frequently misses the queried small targets (e.g., scooter, baby carriage), or recalls only a subset of the relevant instances when multiple objects co-occur.
In contrast, LazyMCoT couples cross-modal attention with a visual expert and further refines the evidence pool through Stage~2 re-querying inside enlarged attention crops.

\section{Conclusion}
We presented \textbf{LazyMCoT}, a training-free framework allocating visual grounding effort based on sample difficulty. Its \textbf{Adaptive Routing} uses first-token statistics and conformal prediction to bypass easy samples. 
For routed samples, \textbf{Collaborative Grounding} combines cross-modal attention and a visual expert to generate precise localized panel displays robust to small or occluded targets. 
Experiments demonstrate LazyMCoT achieves competitive accuracy, surpasses recent training-based methods, and reduces inference latency. This selective grounding paradigm offers a practical recipe for efficient visual reasoning with frozen VLMs.

\bibliography{aaai2027}

\clearpage
\section*{Focus When Necessary: Adaptive Routing and Collaborative Grounding for Training-Free Visual Grounding}
\section*{Appendix}

\vspace{1em}
\hrule
\vspace{1em}
\vspace{\baselineskip}

\renewcommand{\thefootnote}{\fnsymbol{footnote}}
\renewcommand{\thesection}{\Alph{section}}
\renewcommand{\thetable}{\Alph{table}}
\renewcommand{\theequation}{\Alph{equation}}
\renewcommand{\thefigure}{\Alph{figure}}

\setcounter{section}{0}
\setcounter{table}{0}
\setcounter{section}{0}
\setcounter{figure}{0}
\setcounter{equation}{0}

\subsection{Content}
This Appendix is organized as follows:
\begin{itemize}
    \item \textbf{Implementation Details.} Hardware and reproducibility settings, VLM backbones and the SAM3 visual expert, the full Adaptive Routing training pipeline, and the Collaborative Grounding hyperparameters.
    \item \textbf{Dataset Information.} Detailed descriptions of the four high-resolution multiple-choice benchmarks used in our experiments, including image source, resolution, task taxonomy, and evaluation metric.
    \item \textbf{More Statistical Features.} Routing score and entropy diagnostic plots for all three VLMs, demonstrating that the proposed first-token statistics are VLM-agnostic.
    \item \textbf{More Qualitative Results.} Additional side-by-side comparisons between HiDe and LazyMCoT, illustrating how Collaborative Grounding produces cleaner Localized Panel Displays for the VLM re-query.
    \item \textbf{Some Inference Cases.} End-to-end inference traces that compare the base VLM, HiDe, and LazyMCoT on some samples, showing the complementary roles of Adaptive Routing and Collaborative Grounding.
\end{itemize}
\bigskip

\section{Implementation Details}
\label{sec:supp_impl}
For each VLM backbone, we conduct experiments on two NVIDIA H20 GPUs.
To ensure reproducibility, we fix the random seeds for all libraries (Python, CUDA, PyTorch, and NumPy) to 2077 during the training process.

\subsection{VLM Backbones and Visual Expert}
LazyMCoT is evaluated on three open-source VLM backbones: Qwen2.5-VL-7B-Instruct~\cite{qwen2_5vl}, Qwen3-VL-8B-Instruct~\cite{qwen3vl}, and InternVL3-8B-Instruct~\cite{internvl3}. 
All backbones remain frozen during inference and we use greedy decoding (\texttt{do\_sample=False}). Input images are resized so that their longest edge is at most $\texttt{maxp}=16{,}384$ pixels to fit the dynamic-resolution encoders. 
We adopt SAM3~\cite{sam3} as the unified visual expert and serve it as a separate FastAPI service that accepts entity texts as prompts and returns boxes, masks, and confidence scores.

\subsection{Adaptive Routing}
\boldparagraph{Feature extraction.}
For each test sample, we run a single forward pass with \texttt{max\_new\_tokens}$=1$ and record the first answer token logits $\mathbf{z}\in\mathbb{R}^{V}$. We then parse the candidate option letters $\mathcal{O}$ from the question text via regular expression so that benchmarks with $K\geq 4$ options are handled uniformly. 
The same forward pass also produces the direct answer letter, which is reused when the router decides to skip Collaborative Grounding.

\boldparagraph{Router training.}
We split the unified routing set $\mathcal{D}_{\mathrm{cal}}$ by the base VLM into ori-correct ($y=0$) and ori-wrong ($y=1$) samples and train a Gradient Boosting Decision Tree~\cite{gbdt} ($g_\theta$) with $300$ estimators, max depth $3$, and learning rate $0.05$ under $5$-fold cross-validation. The out-of-fold (OOF) predicted probability $\hat p(\mathbf{x})$ is mapped to the routing score $s(x)=\log\hat p/(1-\hat p)$. 
Unless otherwise stated we adopt $\alpha=0$ for the strictest must-recall guarantee. The trained router is serialized once into a JSON report.

\subsection{Collaborative Grounding}
\begin{figure*}
   \centering
   \includegraphics[width=1.0\linewidth]{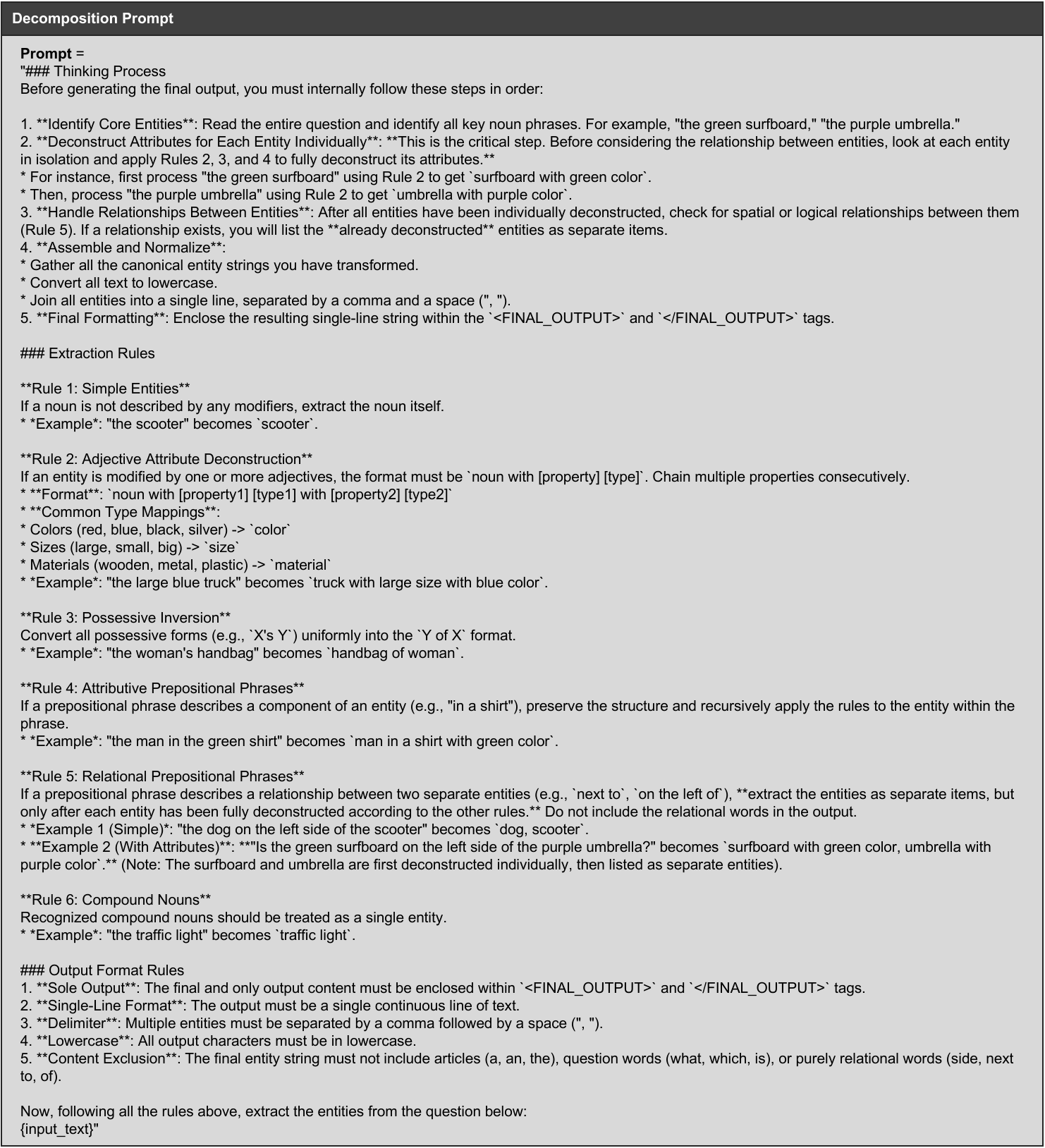}
   \caption{Decomposition prompt template}
   \label{fig:prompt}
\end{figure*}

\boldparagraph{Entity decomposition.}
The base VLM is prompted with a rule-based template~\cite{hide} to decompose the question $Q$ into a list of canonical entities $\mathcal{E}=\{e_1,\dots,e_M\}$. 
The decomposition prompt is shwon as Fig.~\ref{fig:prompt}.

\boldparagraph{Visual expert branch.}
Each entity in $\mathcal{E}$ is sent to SAM3 as an independent text prompt. We retain at most $k=10$ boxes per entity and apply cross-entity NMS with IoU threshold $0.7$ to remove duplicate detections of the same instance.

\boldparagraph{Attention branch.}
For each entity token we reshape the attention to the visual-token grid, apply a Gaussian blur with $\sigma=3$, and linearly normalize the map to $[0,1]$. Connected components above the relative threshold $\tau=0.5$ are converted into the attention boxes $\mathcal{B}_{\mathrm{att}}$.

\boldparagraph{Per-VLM attention layers.}
The cross-modal attention consumed by the attention branch is read from the self-attention modules of the decoder, averaged over all attention heads. Since different VLMs expose their most discriminative grounding signal at different depths, we select the attention source per VLM. For Qwen2.5-VL-7B and InternVL3-8B (both 28 decoder layers), we extract the attention of a single mid-level decoder layer, namely the $15$-th layer ($1$-indexed). For Qwen3-VL-8B (36 decoder layers), we instead aggregate the attention of all decoder layers, because its DeepStack multi-layer visual injection distributes grounding cues across depths so a single layer is insufficient. When more than one layer is used, the per-layer maps are first averaged over heads and then averaged across the selected layers to form the aggregated saliency map $\mathcal{A}(I)$.

\section{Dataset Information}
\label{sec:supp_data}

We evaluate LazyMCoT on four challenging high-resolution multiple-choice benchmarks. All benchmarks adopt accuracy as the primary evaluation metric.

\subsection{V$^\ast$ Bench}
V$^\ast$ Bench~\cite{V-Star} is introduced together with the V$^\ast$ visual search algorithm to evaluate the ability of multimodal LLMs to localize and reason over small or visually inconspicuous targets in high-resolution natural scenes. The benchmark contains $191$ images with an average resolution of $2246\times 1582$, sourced primarily from the SA-1B collection. Each image is paired with a multiple-choice question that targets one of two abilities: (i) \textit{Direct Attribute Recognition} (Att.), which asks about color, material, shape, or other intrinsic attributes of a small object, and (ii) \textit{Spatial Relationship Reasoning} (Spa.), which requires inferring relative positions between two or more objects. Because the queried targets occupy only a tiny fraction of the image, V$^\ast$ Bench is widely used as a stress test for fine-grained visual grounding.

\subsection{HR-Bench-4K and HR-Bench-8K}
HR-Bench~\cite{HR-Bench} is the first benchmark explicitly designed to evaluate MLLMs at $4$K and $8$K resolutions, addressing the gap that previous benchmarks rarely exceed $2$K. HR-Bench provides two splits, HR-Bench-4K and HR-Bench-8K, each consisting of high-resolution images that are evenly partitioned into two task types: \textit{Single-Instance Perception} (Sin.), where the question concerns a single fine-grained object that may be small or inconspicuous, and \textit{Cross-Instance Perception} (Cro.), where the question requires comparing or relating multiple instances across the image. Human accuracy is $87\%$~\cite{HR-Bench}, highlighting the difficulty of high-resolution understanding and making HR-Bench an ideal testbed for visual grounding.

\subsection{TreeBench}
TreeBench~\cite{TreeBench} is a recent benchmark that probes both fine-grained perception and high-order reasoning under traceable visual evidence. It is constructed by sampling $1{,}000$ object-dense images from SA-1B and, after a three-stage manual quality control by eight LMM experts, retains $405$ challenging multiple-choice VQA pairs. Each question is annotated with both the answer and the corresponding ground-truth bounding boxes, ensuring that the evaluation rewards genuine localization rather than language priors. The benchmark is organized into ten fine-grained categories grouped under two competencies: \textbf{Perception} (\textit{Attributes}, \textit{Material}, \textit{Physical State}, \textit{Object Retrieval}, \textit{OCR}) and \textbf{Reasoning} (\textit{Perspective Transformation}, \textit{Ordering}, \textit{Contact \& Occlusion}, \textit{Spatial Containment}, \textit{Comparison}).

\section{More statistical features}
\label{sec:supp_stat}
\begin{figure}[t]
    \centering
    \subfloat[Qwen2.5-VL-7B-Instruct]{
        \includegraphics[width=0.95\linewidth]{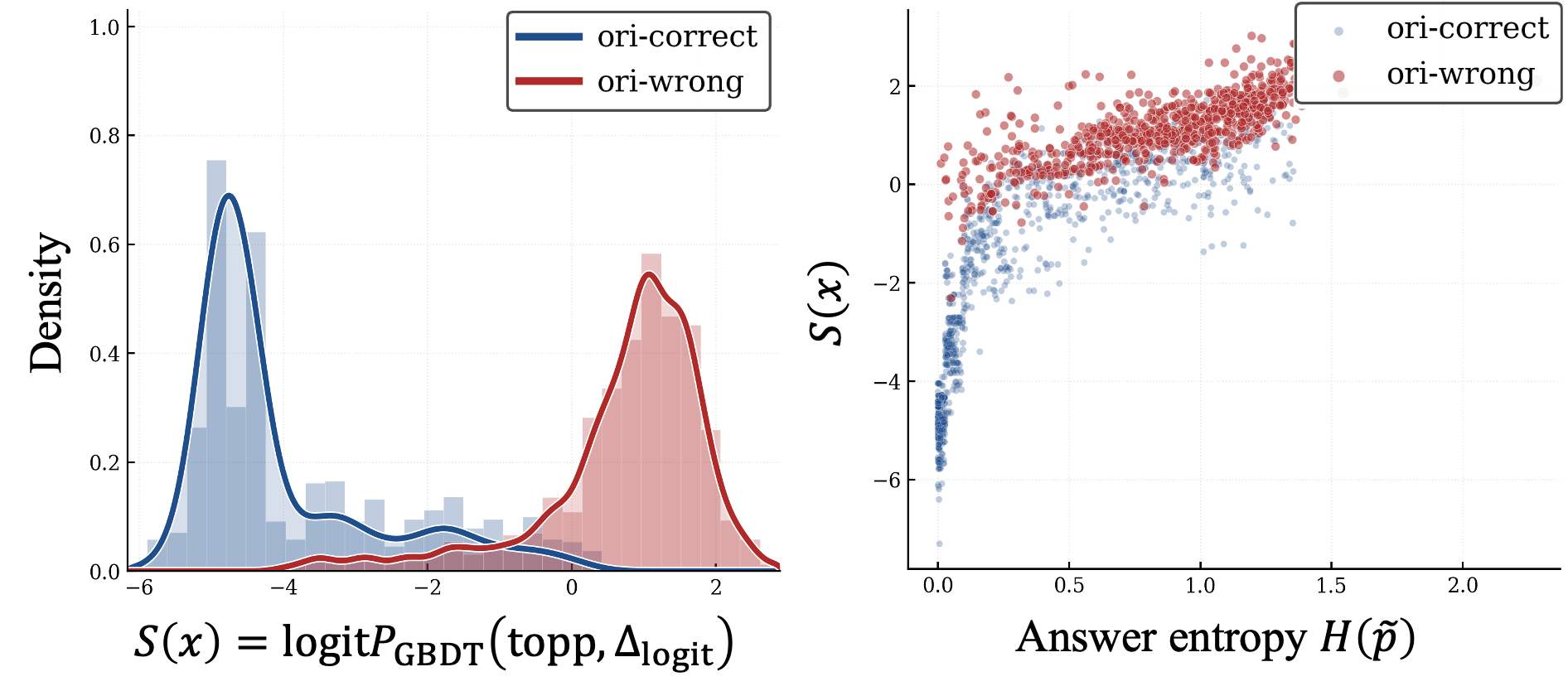}%
        \label{fig:stat_appendix_a}
    }\\
    \subfloat[InternVL3-8B-Instruct]{
        \includegraphics[width=0.95\linewidth]{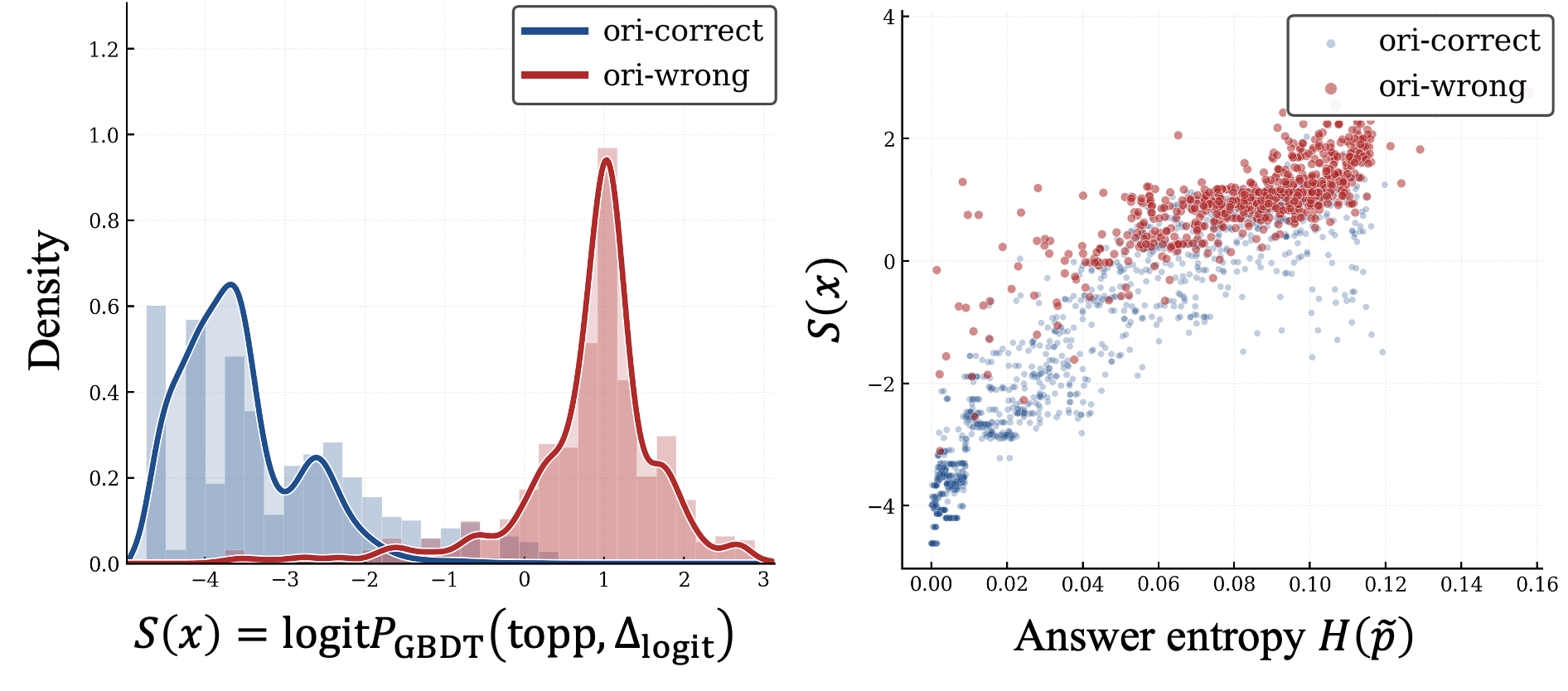}%
        \label{fig:stat_appendix_b}
    }\\
    \subfloat[Qwen3-VL-8B-Instruct]{
        \includegraphics[width=0.95\linewidth]{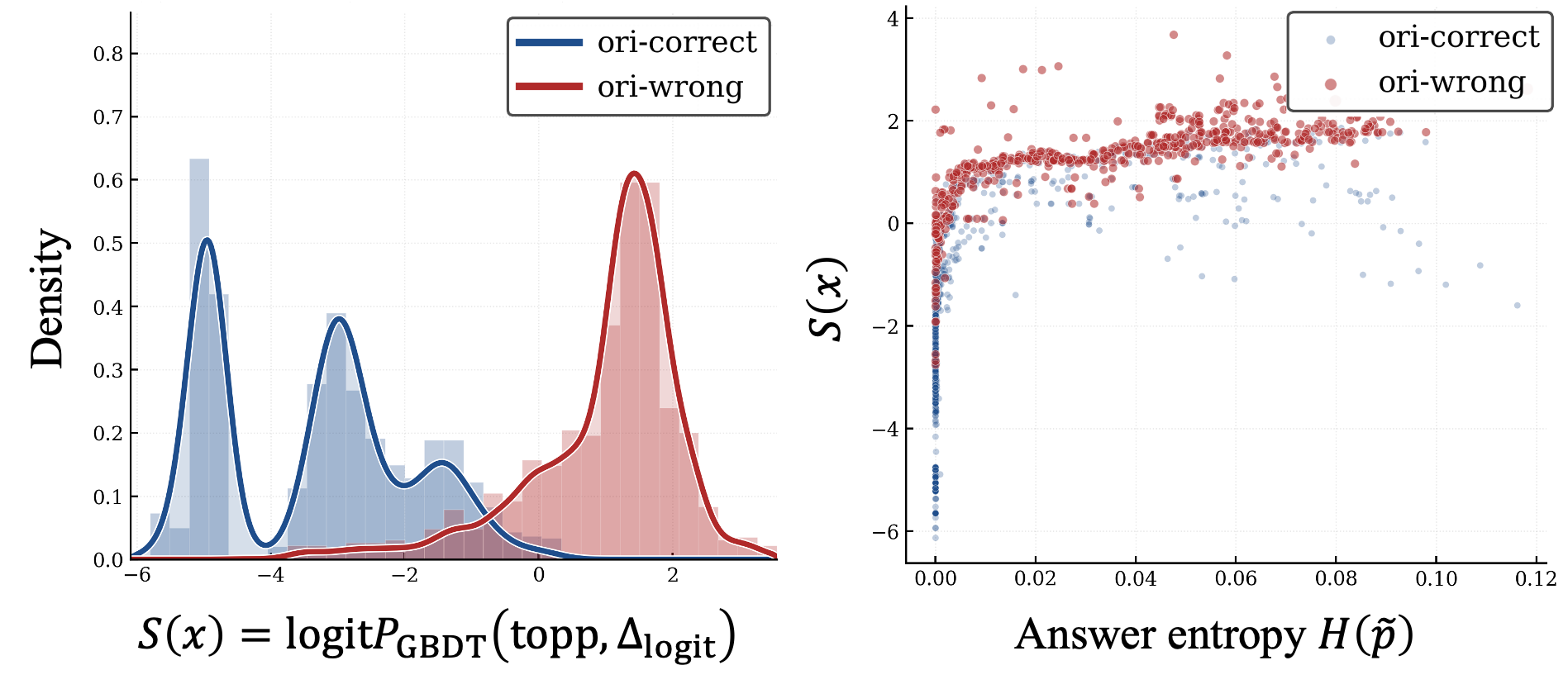}%
        \label{fig:stat_appendix_c}
    }
    \caption{\textbf{Statistical features for sample routing across VLM backbones.} For each backbone, the left panel plots the routing score $s(x)$ distribution, and the right panel plots $s(x)$ versus the option entropy $H(\tilde p)$. The same separability pattern holds across all backbones, validating that the proposed first-token statistics are VLM-agnostic.}
    \label{fig:stat_appendix}
\end{figure}

To verify that the proposed statistics in Sec.~3 generalize beyond a single VLM backbone, we report the same two diagnostic plots on all three evaluated VLMs in Fig.~\ref{fig:stat_appendix}. 
The left panel of each row shows the distribution of the GBDT routing score $s(x)$ for ori-correct and ori-wrong samples, while the right panel shows the joint distribution of $s(x)$ versus the option entropy $H(\tilde p)$. 
Across Qwen2.5-VL-7B, InternVL3-8B, and Qwen3-VL-8B, the routing score consistently produces well-separated modes between the two classes, and is monotonically correlated with answer entropy with ori-wrong samples concentrated in the high-score, high-entropy region. 
This consistency confirms that the proposed first-token statistics are not VLM-specific artifacts, and that the same lightweight router can be calibrated for any frozen VLM without architectural assumptions.

\section{More qualitative results}
\label{sec:supp_qualitative}
\begin{figure*}
   \centering
   \includegraphics[width=1.0\linewidth]{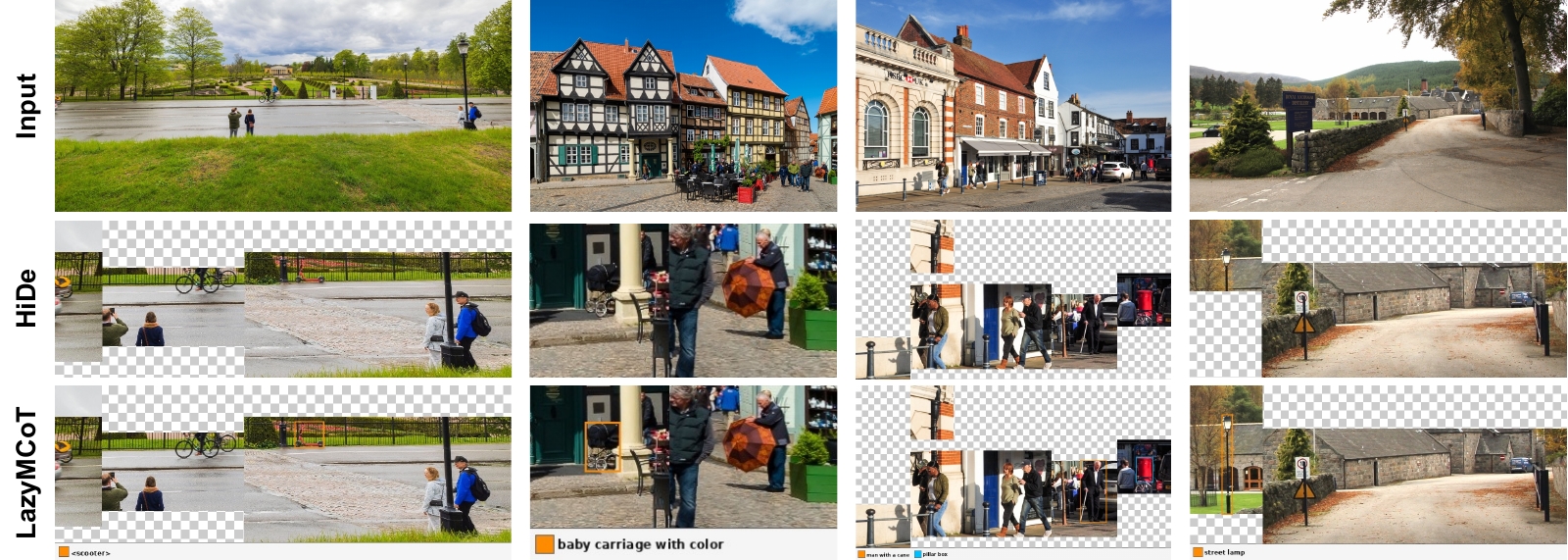}
   \caption{Additional qualitative comparison on V$^\ast$ Bench.}
   \label{fig:vstar}
\end{figure*}

\begin{figure*}
   \centering
   \includegraphics[width=1.0\linewidth]{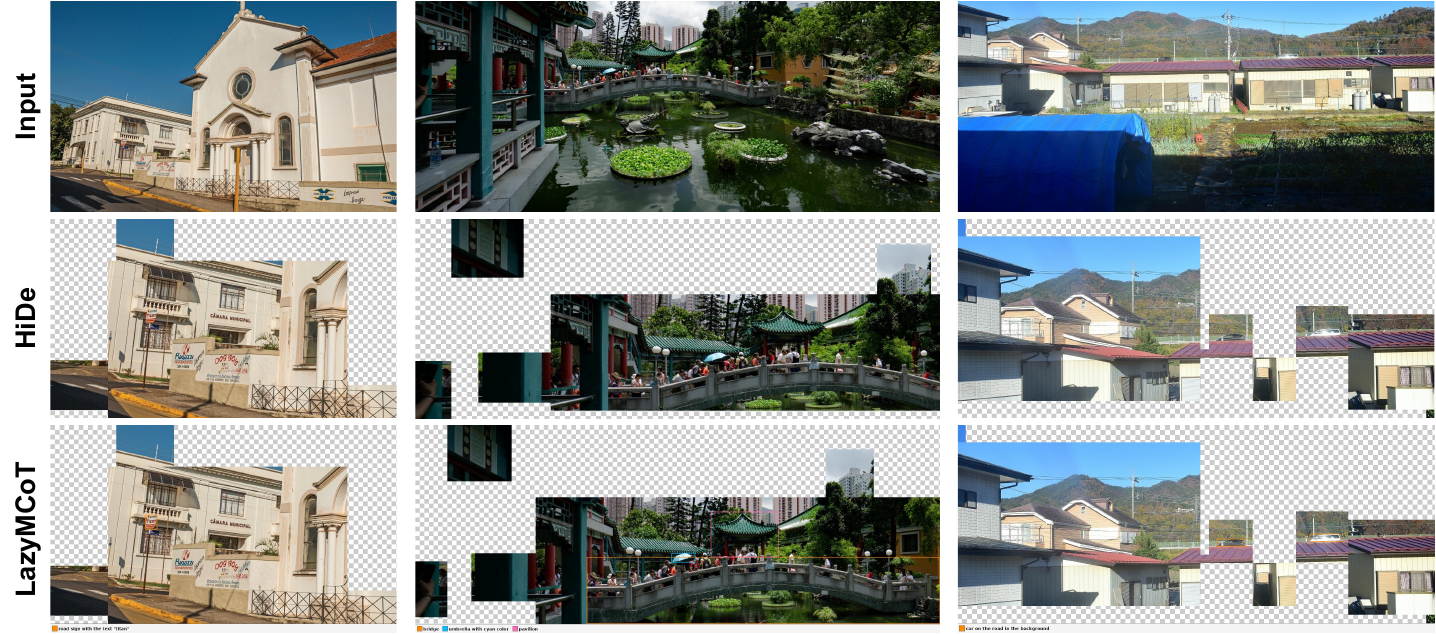}
   \caption{Additional qualitative comparison on TreeBench.}
   \label{fig:treebench}
\end{figure*}

\begin{figure*}
   \centering
   \includegraphics[width=1.0\linewidth]{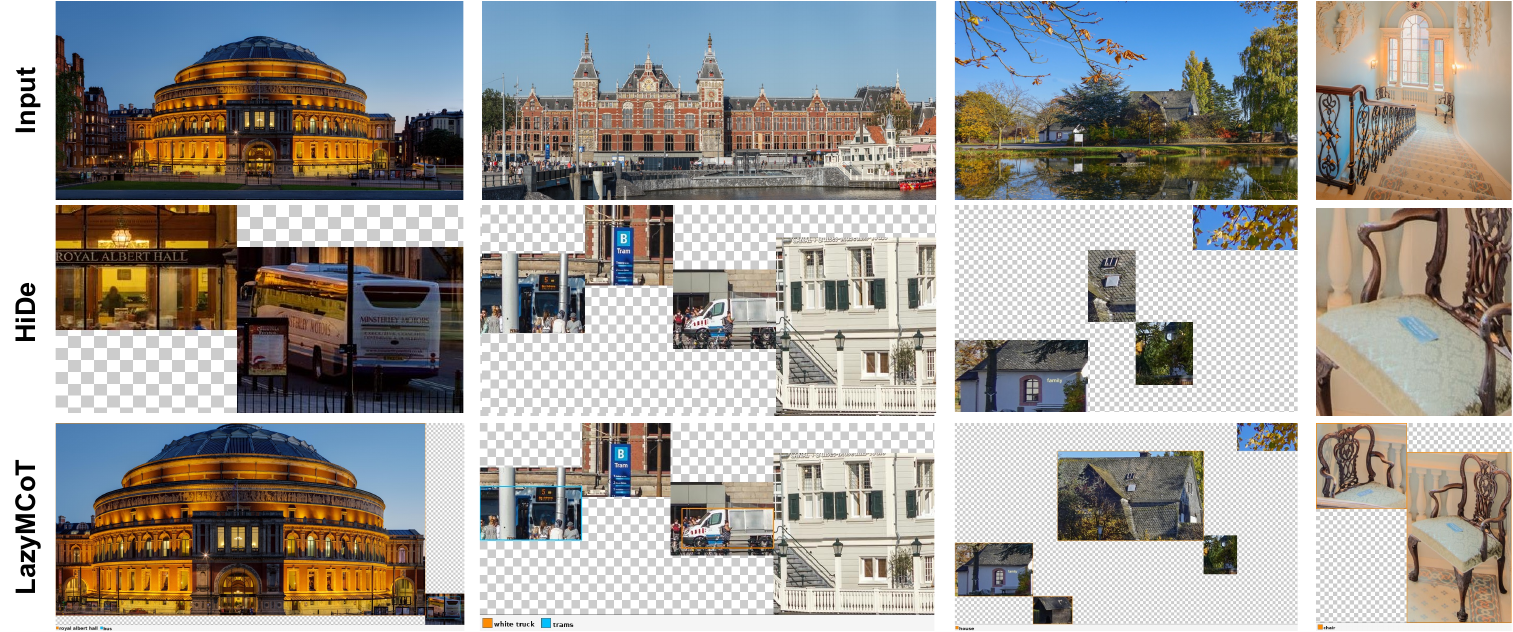}
   \caption{Additional qualitative comparison on HR-Bench.}
   \label{fig:hr-bench} 
\end{figure*}

We provide additional qualitative comparisons on the three high-resolution benchmarks to illustrate how Collaborative Grounding helps the base VLM recover the correct answer on hard samples.
Fig.~\ref{fig:vstar}, Fig.~\ref{fig:treebench}, and Fig.~\ref{fig:hr-bench} respectively present cases drawn from V$^\ast$ Bench, TreeBench, and HR-Bench. 
For each case we show, from left to right, the original image with the question, the Localized Panel Display (LPD) generated by HiDe, and the LPD produced by our LazyMCoT, accompanied by the predicted answers. 
Across all three benchmarks, HiDe frequently misses the queried small or co-occurring targets because its attention-only cropping is sensitive to noise, while LazyMCoT couples cross-modal attention with a SAM3 visual expert and further refines the evidence pool through Stage~2 re-querying. 
The resulting LPDs cover the question-relevant entities more completely and are decorated with per-entity color borders and textual legends, which guide the VLM to faithfully read off the correct answer in a single re-query.

\section{Some Inference Cases}
\label{sec:supp_case}
\begin{figure*}
   \centering
   \includegraphics[width=1.0\linewidth]{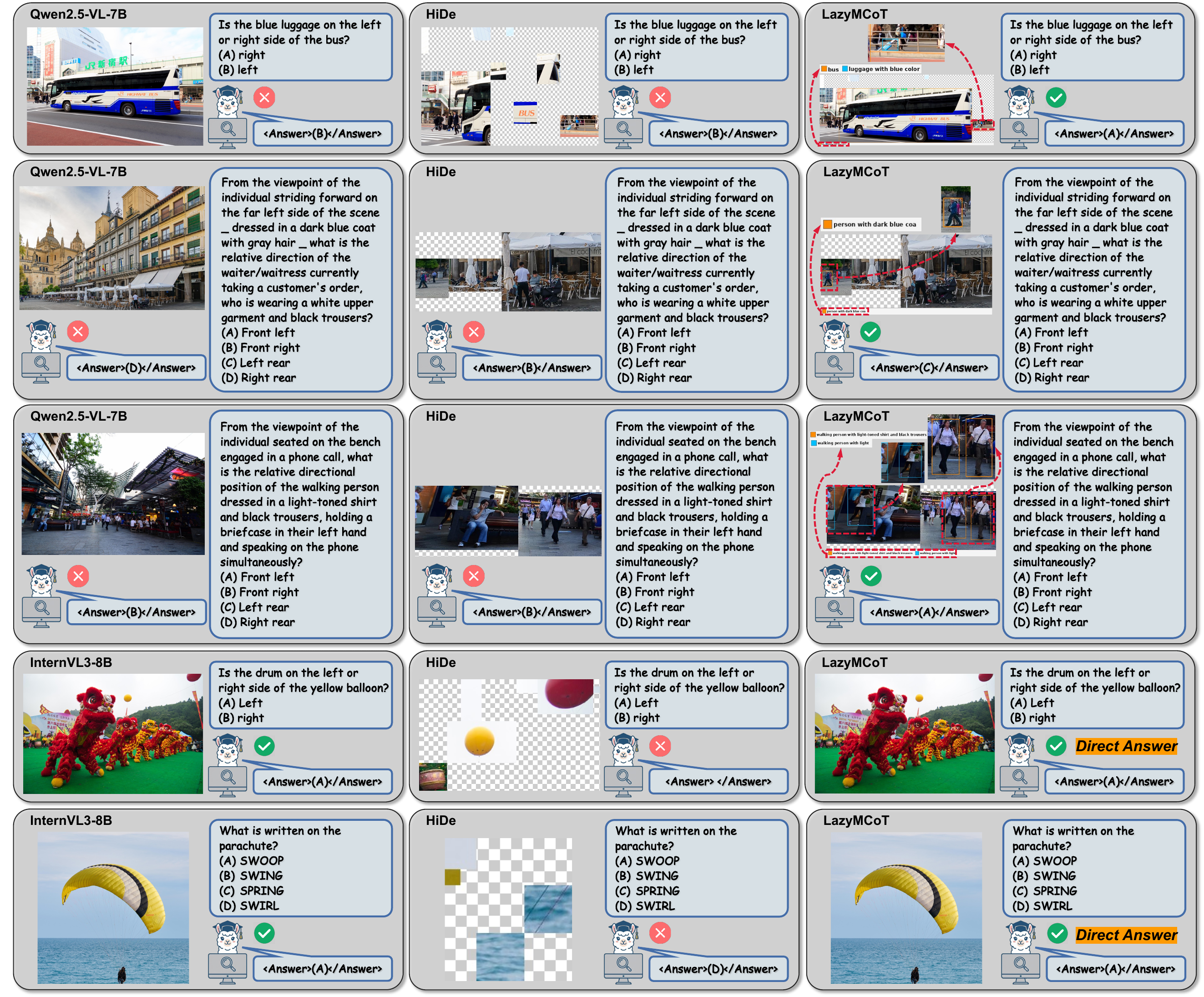}
   \caption{\textbf{End-to-end inference cases comparing the base VLM, HiDe, and LazyMCoT.} The top three rows show hard samples where Adaptive Routing triggers Collaborative Grounding and LazyMCoT produces a clean LPD that recovers the correct answer. The bottom two rows show easy samples where the router emits a \textbf{Direct Answer} and bypasses grounding.}
   \label{fig:case_appendix}
\end{figure*}

Fig.~\ref{fig:case_appendix} presents end-to-end inference traces of LazyMCoT alongside the base VLM and HiDe on five representative samples drawn from V$^\ast$ Bench and HR-Bench. 
Each row shows, from left to right, the prediction of the original VLM on the raw image, the prediction of HiDe on its attention-only LPD, and the prediction of LazyMCoT on its Collaborative Grounding output. 
The first three rows correspond to hard samples on which the base Qwen2.5-VL-7B fails. Adaptive Routing identifies them as uncertain and dispatches them to Collaborative Grounding, which produces LPDs in which the queried entities are precisely highlighted by SAM3 colored boxes and textual legends. 
With this faithful evidence, the same VLM corrects its answer in a single re-query. 
The last two rows correspond to easy samples on which the base InternVL3-8B already answers correctly. 
Here HiDe's forced grounding truncates the global context and misleads the VLM into wrong predictions, whereas Adaptive Routing tags the samples as confident and emits the direct answer immediately, as marked by the \textbf{Direct Answer} flag in the rightmost column. 
Together, these traces illustrate the complementary roles of the two components, Collaborative Grounding rescues hard samples that the VLM cannot solve alone, while Adaptive Routing prevents Collaborative Grounding from interfering with samples that are already well solved.

\end{document}